# Modeling bank performance: A novel fuzzy two-stage DEA approach

Mohammad Izadikhah

*Department of Mathematics, College of Science, Arak Branch, Islamic Azad University, Arak, Iran; Emails: m-izadikhah@iau-arak.ac.ir; m_izadikhah@yahoo.com;*

**Abstract**:

Evaluating the banks' performance has always been of interest due to their crucial role in the economic development of each country. Data envelopment analysis (DEA) has been widely used for measuring the performance of bank branches. In the conventional DEA approach, decision making units (DMUs) are regarded as black boxes that transform sets of inputs into sets of outputs without considering the internal interactions taking place within each DMU. Two-stage DEA models are designed to overcome this shortfall. Thus, this paper presented a new two-stage DEA model based on a modification on Enhanced Russell Model. On the other hand, in many situations, such as in a manufacturing system, a production process or a service system, inputs, intermediates and outputs can be given as a fuzzy variable. The main aim of this paper is to build and present a new fuzzy two-stage DEA model for measuring the efficiency of 15 branches of Melli bank in Hamedan province.

Subject classification codes: Mathematics & Statistics; Applied Mathematics; Mathematical Modeling; Non-Linear Systems

**1 Introduction**

Banks and financial and credit institutions play a very important role in the economic development of each country. Currently, due to the significant growing the number of banks and financial and credit institutions in Iran and due to the privatization process of state banks and the conversion of credit cooperatives and credit institutions to the bank, their performance evaluation has become very important. One of the well-known methods in assessing the performance of firms is data envelopment analysis that was developed by Charnes et al. [1] as CCR model, see ([2-6]; [7]) for more details. CCR model is a radial model. Radial models have some disadvantages like failure to recognize weak efficient DMUs, see [8,9]. Another kind of DEA models are non-radial DEA models [10]. One of the important non-radial DEA model is Enhanced Russell Model (ERM model) that proposed by [11]. This model has some useful properties. One of them is ability to recognize the weak efficient DMUs. On the other hand this model has some disadvantages like failure to rank efficient DMUs. Izadikhah et al. [12] proposed a modified version of ERM model that enables ERM model to rank efficient DMUs. Our proposed DEA methodology is an extension of their model that considers internal structures of DMUs. DEA has been widely used for assessing the performance of banks. Chortareas et al. [13] presented a good survey in this topic. Also, the readers are referred to [14] for a complete review for applications of DEA in industries.

Typically, a single stage production process is assumed to transform inputs to final outputs and is treated as a black box. In contrast to the black-box approach, real-world production systems often have network structure [15]. There is an increasing literature body that is devoted to efficiency assessment in multistage production processes. Castelli et al. [16] provided a comprehensive review of models and methods developed for different multi-stage production structures. In recent years, many researchers studied various DEA models for evaluating efficiencies of two-stage systems especially for evaluating efficiencies of banking systems. Firstly, Seiford and Zhu [17] presented the first two-stage DEA model to evaluate the marketability and profitability of the U.S. commercial banks. For an in-depth review in multi stage DEA model, see ([9,18-21]). In this study, we propose a new two-stage DEA model based on the modified ERM model. However, in many situations, such as in a manufacturing system, a production process or a service system, inputs and outputs are volatile and complex so that it is difficult to measure them in an accurate way. Instead, the data can be given as a fuzzy variable. The concept of fuzzy theory was initialized in Zadeh [22]. After that many fuzzy approaches have been introduced in the DEA literature. Sengupta [23] applied principle of fuzzy set theory to introduce fuzziness in the objective function and the right-hand side vector of the conventional DEA model and developed the tolerance approach that was one of the first fuzzy DEA models. Many authors have combined DEA and Fuzzy modeling to study the efficiency of banking systems, see [24] and for a review on fuzzy DEA modelling see ([25-31]; [32]; [33,34]). Conventional DEA needs accurate measurement of inputs and outputs. However, the values of the input and output data in banking systems are sometimes imprecise or uncertain and since the quantity of some of our data in this paper was not known exactly the data was stated as fuzzy data. Thus, we extend our proposed two-stage DEA model in fuzzy environment. Then we present a method for solving the proposed two-stage fuzzy DEA model based on the concept of alpha cut and possibility approach. Also for the purpose of final ranking and calculating the overall and stage efficiencies, we present a stochastic closeness coefficient. This coefficient is very useful and integrates all results of various values of $\alpha$. The objective of this paper is to present a new two-stage fuzzy DEA model based on a modified version of enhanced Russell measure model to measure the performance of 15 branches of Melli bank in Hamedan province in the time period 2015-2016. In the proposed case study, the data are known as triangular fuzzy numbers.

The main contributions of this paper are as follows: This paper presents a new two-stage DEA model based on a recent modified version of ERM model. Also, this paper presents a new two-stage fuzzy DEA model based on the modified ERM. Our model uses the concept of $\alpha$-cut and possibility approach to defuzzification. Also for the purpose of calculating the overall and stage efficiencies this paper proposes a stochastic closeness coefficient. This coefficient removes the difficulty of various ranking by various values of $\alpha$. The proposed methodology applies to evaluate the efficiencies of 15 branches of Melli bank in Hamedan province.

This paper unfolds as follows: Section 2 briefly reviews the possibility approach. Section 3 proposes our new DEA methodology. In Section 4, a case study is presented and final conclusion is appeared in Section 5.

## 2 Preliminaries

Zadeh [22] presented the concept of possibility approach in terms of fuzzy set theory. Now, we review the definition of possibility space, ([22,35]).

**Definition 1:** (Possibility space) Let $\Theta$ be a nonempty set, and P the power set of $\Theta$. Each element in P is called an event. To present an axiomatic definition of possibility, it is necessary to assign to each event A, a number $\pi(A)$ which indicates the possibility that A will occur. Then the triplet $(\Theta, p, \pi)$ is called a possibility space.

**Definition 2:** Let A be a fuzzy variable defined on a possibility space $(\Theta, p, \pi)$. The membership of this variable introduced by Zadeh is as follows:

$$\mu_A(s) = \pi(\theta_i \in \Theta_i | A(\theta_i) = s) = \sup_{\theta_i \in \Theta_i} \{\pi(\theta_i) | A(\theta_i) = s\}, \quad \forall s \in R$$

**Definition 3:** Let $(\Theta, p, \pi)$ be a possibility space such that $\Theta = \Theta_1 \times \ldots \times \Theta_n$, therefore, for any set A we have

$$\pi(A) = \sup_{\theta_i \in \Theta_i} \{\pi_i(A_i) \mid A = A_1 \times L \times A_n, A_i \in p\}$$

**Definition 4:** We denote an $\alpha$-cut of fuzzy number $A$ by $A_\alpha$ which is defined as follows:

$$A_\alpha = \{x \mid A(x) \geq \alpha\}$$

An $\alpha$-cut of $A$ can be stated as $A_\alpha = [A_L^{-1}(\alpha), A_U^{-1}(\alpha)] = \left[ [A]_\alpha^L, [A]_\alpha^U \right]$ for all $\alpha \in [0,1]$.

Considering the fuzzy theory, there is a lemma that can be very useful to interpret the possibility function. Now, this lemma is represented.

**Lemma 1:** Let $A_1, \ldots, A_n$ be normal and convex fuzzy variables. Then, for any given possibility levels $\varepsilon_1$, $\varepsilon_2$ and $\varepsilon_3$ ($0 \leq \varepsilon_i \leq 1$) we have

(i): $\pi(A_1 + \ldots + A_n \leq a) \geq \varepsilon_1$ if and only if $[A_1]_{\varepsilon_1}^L + \ldots + [A_n]_{\varepsilon_1}^L \leq a$

(ii): $\pi(A_1 + \ldots + A_n \geq a) \geq \varepsilon_2$ if and only if $[A_1]_{\varepsilon_2}^U + \ldots + [A_n]_{\varepsilon_2}^U \geq a$

(iii): $\pi(A_1 + \ldots + A_n = a) \geq \varepsilon_3$ if and only if $[A_1]_{\varepsilon_3}^L + \ldots + [A_n]_{\varepsilon_3}^L \leq a$ and $[A_1]_{\varepsilon_3}^U + \ldots + [A_n]_{\varepsilon_3}^U \geq a$

Where $\left[A_j\right]_{\varepsilon_i}^L$ and $\left[A_j\right]_{\varepsilon_i}^U$ are the lower and upper bounds of the $\varepsilon_i$-level set of $A_j$ $(j=1,\ldots,n)$. The above lemma is very useful to defuzzification of the fuzzy DEA model's constrains.

## 3 Proposed Methodology

In this section, we first present our proposed two-stage DEA model, and then we bring our new fuzzy two-stage DEA model alongside the procedure for solving its program.

### 3.1 Proposed two-stage DEA model

Let's consider DMUs have an internal production structure. So, in this section we extend black-box production structure and performance measures to a two-stage production process. Here, assume that there are *n* DMUs *(j=1,...,n)* consisting of *two* divisions and $m_1$ and $s_1$ are numbers of inputs and outputs of the first Division and $m_2$ and $s_2$ are numbers of inputs and outputs of the second Division, respectively. Also, $x_{ij}^1$, $(i=1,...,m_1)$ denotes input that is consumed by the first stage of $DMU_j$ entirely, and $y_{rj}^1$, $(r=1,...,s_1)$ denotes output that is produced by first stage of $DMU_j$ directly. And, $x_{ij}^2$, $(i=1,...,m_2)$ denotes input that is consumed by the second stage of $DMU_j$, entirely; $y_{rj}^2$, $(r=1,...,s_2)$ denotes output that is produced by second stage of $DMU_j$, directly. $z_{fj}$, $(f=1,...,F)$ shows intermediate products from the first division to the second division. We propose a new non-radial two-stage DEA model. Tone and Tsutsui [36] proposed a non-radial network DEA model. Their model evaluated the stage (divisional) efficiencies along with overall efficiency of decision making units. Izadikhah et al. [12] presented a modified version of Enhanced Russel Measure of efficiency (ERM) model that could rank all DMUs. So, in this paper inspired from the idea of [36] and [12] we propose a new two-stage DEA model that carries the properties of two above models. Therefore, our proposed two-stage model under variable returns-to-scale (VRS) case is presented as follows:

$$R_p^* = \min \frac{w_1 \left\{ \frac{1}{m_1} \sum_{i=1}^{m_1} \theta_i^1 \right\} + w_2 \left\{ \frac{1}{m_2} \sum_{i=1}^{m_2} \theta_i^2 \right\}}{w_1 \left\{ \frac{1}{s_1} \sum_{r=1}^{s_1} \phi_r^1 \right\} + w_2 \left\{ \frac{1}{s_2} \sum_{r=1}^{s_2} \phi_r^2 \right\}}$$

*s.t.*

$$\sum_{\substack{j=1\\j\neq p}}^{n} \lambda_j^1 x_{ij}^1 \leq \theta_i^1 x_{ip}^1; \quad i=1,..,m_1; \qquad \sum_{\substack{j=1\\j\neq p}}^{n} \lambda_j^2 x_{ij}^2 \leq \theta_i^2 x_{ip}^2; \quad i=1,..,m_2;$$

$$\sum_{\substack{j=1\\j\neq o}}^{n} \lambda_j^1 y_{rj}^1 \geq \phi_r^1 y_{rp}^1; \quad r=1,..,s_1; \qquad \sum_{\substack{j=1\\j\neq o}}^{n} \lambda_j^2 y_{rj}^2 \geq \phi_r^2 y_{rp}^2; \quad r=1,..,s_2;$$

$$\sum_{\substack{j=1\\j\neq o}}^{n} \lambda_j^1 z_{fj} \leq \sum_{\substack{j=1\\j\neq o}}^{n} \lambda_j^2 z_{fj}; \quad f=1,..,F; \tag{1}$$

$$\sum_{\substack{j=1\\j\neq p}}^{n} \lambda_j^1 = 1; \qquad \sum_{\substack{j=1\\j\neq p}}^{n} \lambda_j^2 = 1;$$

$$\theta_i^1 - 1 \leq M\delta^1; \quad i=1,..,m_1; \qquad \theta_i^2 - 1 \leq M\delta^2; \quad i=1,..,m_2;$$

$$-\theta_i^1 + 1 \leq M(1-\delta^1); \quad i=1,...,m_1; \qquad -\theta_i^2 + 1 \leq M(1-\delta^2); \quad i=1,...,m_2;$$

$$-\phi_r^1 + 1 \leq M\delta^1; \quad r=1,...,s_1; \qquad -\phi_r^2 + 1 \leq M\delta^2; \quad r=1,...,s_2;$$

$$\phi_r^1 - 1 \leq M(1-\delta^1); \quad r=1,...,s_1; \qquad \phi_r^2 - 1 \leq M(1-\delta^2); \quad r=1,...,s_2;$$

$$\delta^1, \delta^2 \in \{0,1\}, \qquad \theta_i^1, \theta_i^2, \lambda_j^1, \lambda_j^2 \geq 0, \quad \forall i; \; \forall j$$

where $\lambda_j^1, \lambda_j^2$ ($\forall j$) are intensity variables for the first and second stage of production process. Moreover, $w_1$ and $w_2$ are weights addressing total preference over the two stages. When stages 1 and 2 have similar importance, $w_1$ and $w_2$ will be equal and they add up to 1. On the other hand, set of constraints for intermediate products can be written as follows:

$$\sum_{\substack{j=1 \\ j \neq o}}^{n} \lambda_j^1 z_{fj} \leq \sum_{\substack{j=1 \\ j \neq o}}^{n} \lambda_j^2 z_{fj}; \quad f=1,...,F;$$

In model (1), binary variables $\delta^1$ and $\delta^2$ guarantee that only one group of two groups of constrains is held:

$$(I): \begin{cases} \theta_i^k \leq 1; & \forall i; \forall k=1,2 \\ \phi_r^k \geq 1; & \forall r; \forall k=1,2 \end{cases} \quad \text{or} \quad (II): \begin{cases} \theta_i^k \geq 1; & \forall i, \forall k=1,2 \\ \phi_r^k \leq 1; & \forall r, \forall k=1,2 \end{cases}$$

If DMU is located inside production possibility set (PPS), thus constrains of group (*I*) will be active. If DMU is located outside PPS, thus constrains of group (*II*) will be active. This point enables our model to rank DMUs (see [12]). $R_p^*$ shows overall efficiency score for $DMU_p$. Based on this model, overall efficiency can be defined as follows:

**Definition 5** (Overall efficiency): $DMU_p$ is said to be an overall efficient DMU if $P_p^* \geq 1$, otherwise it is inefficient.

As mentioned before, our model calculates overall efficiency scores and it is able to present a complete ranking that is difficulty of existing models. The $P_p^* \geq 1$ implies that DMU under evaluation is overall efficient and shows rank of DMU. In order to check the performance of each stage, we need to define stage efficiency. However, we define stage efficiency as follows:

$$R_p^1 = \frac{\frac{1}{m_1}\sum_{i=1}^{m_1}\theta_i^*}{\frac{1}{s_1}\sum_{r=1}^{s_1}\phi_r^*} \qquad R_p^2 = \frac{\frac{1}{m_2}\sum_{i=1}^{m_2}\theta_i^*}{\frac{1}{s_2}\sum_{r=1}^{s_2}\phi_r^*}$$

where $\theta_i^*$ and $\phi_r^*$ are appeared in optimum solution of model (1). $R_p^k$ (k=1, 2) shows $k^{th}$-stage efficiency score for $DMU_p$ and based on this model stage efficiency can be defined as follows:

**Definition 6** (Stage efficiency): $DMU_p$ is $k^{th}$-stage efficient DMU if $P_p^k \geq 1$, otherwise it is inefficient.

### 3.2 A new Fuzzy two-stage modified ERM model

The classic DEA models can only be used for cases where the data are precisely measured while in real-world situations, the observed values of the input and output data are sometimes inexact, incomplete, vague or ambiguous. These kinds of uncertainty data can be represented as linguistic variables characterized by fuzzy numbers for reflecting a kind of general sense or experience of experts. The concept of fuzzy set theory was first developed by [22] to deal with the issue of uncertainty in systems modeling. Fuzzy DEA is a powerful tool for evaluating the performance of DMUs in uncertainty environments. In this section, we propose a new fuzzy DEA model for evaluating a set of DMUs with fuzzy inputs, intermediates and outputs. Hence, we extend model (1) to a fuzzy model.

### 3.2.1 Justification of the fuzzy model

Let the evaluation of efficiency of a homogeneous set of $n$ DMUs (DMU$_j$; $j = 1,...,n$) is to be assessed where each DMU consists of *two* divisions. Also, $\tilde{x}_{ij}^1$, $(i = 1, ..., m_1)$ denotes $m_1$ fuzzy inputs that are consumed by the first stage of $DMU_j$ entirely, and $y_{rj}^1, (r = 1, ..., s_1)$ denotes $s_1$ fuzzy outputs that are produced by first stage of $DMU_j$ directly. And, $\tilde{x}_{ij}^2$, $(i = 1, ..., m_2)$ denotes $m_2$ fuzzy inputs that are consumed by the second stage of $DMU_j$, entirely; $y_{rj}^2, (r = 1, ..., s_2)$ denotes $s_2$ fuzzy outputs that are produced by second stage of $DMU_j$, directly. $\tilde{z}_{fj}, (f = 1, ..., F)$ shows F fuzzy intermediate products from the first division to the second division. The proposed fuzzy two-stage DEA model for calculating the overall efficiency of $DMU_p$ is as follows:

$$R_p^* = \min \frac{w_1 \left\{\frac{1}{m_1}\sum_{i=1}^{m_1}\theta_i^1\right\} + w_2 \left\{\frac{1}{m_2}\sum_{i=1}^{m_2}\theta_i^2\right\}}{w_1 \left\{\frac{1}{s_1}\sum_{r=1}^{s_1}\varphi_r^1\right\} + w_2 \left\{\frac{1}{s_2}\sum_{r=1}^{s_2}\varphi_r^2\right\}}$$

s.t.

$$\sum_{\substack{j=1\\j\neq p}}^{n} \lambda_j^1 \tilde{x}_{ij}^1 \leq \theta_i^1 \tilde{x}_{ip}^1; \quad i = 1, \dots, m_1$$

$$\sum_{\substack{j=1\\j\neq p}}^{n} \lambda_j^2 \tilde{x}_{ij}^2 \leq \theta_i^2 \tilde{x}_{ip}^2; \quad i = 1, \dots, m_2$$

$$\sum_{\substack{j=1\\j\neq p}}^{n} \lambda_j^1 \tilde{y}_{rj}^1 \geq \varphi_r^1 \tilde{y}_{rp}^1; \quad r = 1, \dots, s_1$$

$$\sum_{\substack{j=1\\j\neq p}}^{n} \lambda_j^2 \tilde{y}_{rj}^2 \geq \varphi_r^2 \tilde{y}_{rp}^2; \quad r = 1, \dots, s_2$$

$$\sum_{\substack{j=1\\j\neq p}}^{n} \lambda_j^1 \tilde{z}_{fj}^1 \leq \sum_{\substack{j=1\\j\neq p}}^{n} \lambda_j^2 \tilde{z}_{fj}^1; \quad f = 1, \dots, F$$

$$\sum_{\substack{j=1\\j\neq p}}^{n} \lambda_j^1 = 1;$$

$$\sum_{\substack{j=1\\j\neq p}}^{n} \lambda_j^2 = 1;$$

$$\theta_i^1 - 1 \leq M\delta^1; \quad i = 1, \dots, m_1;$$
$$-\theta_i^1 + 1 \leq M(1 - \delta^1); \quad i = 1, \dots, m_1;$$
$$\theta_i^2 - 1 \leq M\delta^2; \quad i = 1, \dots, m_2;$$
$$-\theta_i^2 + 1 \leq M(1 - \delta^2); \quad i = 1, \dots, m_2;$$
$$-\varphi_r^1 + 1 \leq M\delta^1; \quad r = 1, \dots, s_1;$$
$$\varphi_r^1 - 1 \leq M(1 - \delta^1); \quad r = 1, \dots, s_1;$$
$$-\varphi_r^2 + 1 \leq M\delta^2; \quad r = 1, \dots, s_2;$$
$$\varphi_r^2 - 1 \leq M(1 - \delta^2); \quad r = 1, \dots, s_2;$$
$$\delta^1, \delta^2 \in \{0,1\}$$
$$\theta_i^1, \varphi_r^1, \lambda_j^1, \theta_i^2, \varphi_r^2, \lambda_j^2 \geq 0; \quad \forall i, r, j$$

(2)

This model is a fuzzy version of model (1) that the fuzzy numbers are incorporated into the model (1). This fuzzy integrated DEA model cannot be solved like a crisp model. It is needed to design a procedure to solve that model.

*3.2.2 Solving Procedure for the proposed fuzzy model*

As it is mentioned before the proposed fuzzy DEA model cannot be solved like a crisp model. So, in order to solve it one can apply a possibility approach formulated in terms of fuzzy set theory proposed by [22]. This procedure converts the fuzzy integrated DEA model to the standard linear programming (LP) by α-cut technique. In this case, each fuzzy coefficient can be viewed as a fuzzy variable and each constraint can be considered as a fuzzy event, see [35]. Using possibility theory, possibilities of fuzzy events (i.e., fuzzy constraints) can be determined. Regarding the proposed model and the concept of possibility space of fuzzy event, some constrains are defined as a crisp value and other constrains are considered as an uncertain. For this reason by introducing the predetermined acceptable levels of possibility for constrains as $\varepsilon_1$, $\varepsilon_2$, $\varepsilon_3$, $\varepsilon_4$ and $\varepsilon_5$. Therefore the proposed model converted as follows:

$$R_p^* = \min \frac{w_1\left\{\frac{1}{m_1}\sum_{i=1}^{m_1}\theta_i^1\right\}+w_2\left\{\frac{1}{m_2}\sum_{i=1}^{m_2}\theta_i^2\right\}}{w_1\left\{\frac{1}{s_1}\sum_{r=1}^{s_1}\phi_r^1\right\}+w_2\left\{\frac{1}{s_2}\sum_{r=1}^{s_2}\phi_r^2\right\}}$$

s.t.

$$\pi\left(\sum_{\substack{j=1\\j\neq p}}^{n}\lambda_j^1\tilde{x}_{ij}^1 - \theta_i^1\tilde{x}_{ip}^1 \leq 0\right) \geq \varepsilon_1; \quad i = 1, \dots, m_1$$

$$\pi\left(\sum_{\substack{j=1\\j\neq p}}^{n}\lambda_j^2\tilde{x}_{ij}^2 - \theta_i^2\tilde{x}_{ip}^2 \leq 0\right) \geq \varepsilon_2; \quad i = 1, \dots, m_2 \quad (3)$$

$$\pi\left(\sum_{\substack{j=1\\j\neq p}}^{n}\lambda_j^1\tilde{y}_{rj}^1 - \varphi_r^1\tilde{y}_{rp}^1 \geq 0\right) \geq \varepsilon_3; \quad r = 1, \dots, s_1$$

$$\pi\left(\sum_{\substack{j=1\\j\neq p}}^{n}\lambda_j^2\tilde{y}_{rj}^2 - \varphi_r^2\tilde{y}_{rp}^2 \geq 0\right) \geq \varepsilon_4; \quad r = 1, \dots, s_2$$

$$\pi\left(\sum_{\substack{j=1\\j\neq p}}^{n}\lambda_j^1\tilde{z}_{fj}^1 - \sum_{\substack{j=1\\j\neq p}}^{n}\lambda_j^2\tilde{z}_{fj}^1 \leq 0\right) \geq \varepsilon_5; \quad f = 1, \dots, F$$

$$\sum_{\substack{j=1\\j\neq p}}^{n}\lambda_j^1 = 1;$$

$$\sum_{\substack{j=1\\j\neq p}}^{n}\lambda_j^2 = 1;$$

$$\begin{aligned}
&\theta_i^1 - 1 \leq M\delta^1; && i = 1, \dots, m_1; \\
&-\theta_i^1 + 1 \leq M(1 - \delta^1); && i = 1, \dots, m_1; \\
&\theta_i^2 - 1 \leq M\delta^2; && i = 1, \dots, m_2; \\
&-\theta_i^2 + 1 \leq M(1 - \delta^2); && i = 1, \dots, m_2; \\
&-\varphi_r^1 + 1 \leq M\delta^1; && r = 1, \dots, s_1; \\
&\varphi_r^1 - 1 \leq M(1 - \delta^1); && r = 1, \dots, s_1; \\
&-\varphi_r^2 + 1 \leq M\delta^2; && r = 1, \dots, s_2; \\
&\varphi_r^2 - 1 \leq M(1 - \delta^2); && r = 1, \dots, s_2;
\end{aligned}$$

$$\delta^1, \delta^2 \in \{0,1\}$$

$$\theta_i^1, \varphi_r^1, \lambda_j^1, \theta_i^2, \varphi_r^2, \lambda_j^2 \geq 0; \qquad \forall i, r, j$$

In model (3) parameters $\varepsilon_1$, $\varepsilon_2$, $\varepsilon_3$, $\varepsilon_4$ and $\varepsilon_5$ are the predefined levels that the related constraints should attain the possibility level. According to Lemma 1, model (3) can be stated as follows:

$$R_p^* = \min \frac{w_1\left\{\frac{1}{m_1}\sum_{i=1}^{m_1}\theta_i^1\right\}+w_2\left\{\frac{1}{m_2}\sum_{i=1}^{m_2}\theta_i^2\right\}}{w_1\left\{\frac{1}{s_1}\sum_{r=1}^{s_1}\phi_r^1\right\}+w_2\left\{\frac{1}{s_2}\sum_{r=1}^{s_2}\phi_r^2\right\}}$$

s.t.

$$\left(\sum_{\substack{j=1\\j\neq p}}^{n}\lambda_j^1\tilde{x}_{ij}^1-\theta_i^1\tilde{x}_{ip}^1\right)_{\varepsilon_1}^{L}\leq 0;\quad i=1,\dots,m_1$$

$$\left(\sum_{\substack{j=1\\j\neq p}}^{n}\lambda_j^2\tilde{x}_{ij}^2-\theta_i^2\tilde{x}_{ip}^2\right)_{\varepsilon_2}^{L}\leq 0;\quad i=1,\dots,m_2 \qquad (4)$$

$$\left(\sum_{\substack{j=1\\j\neq p}}^{n}\lambda_j^1\tilde{y}_{rj}^1-\varphi_r^1\tilde{y}_{rp}^1\right)_{\varepsilon_3}^{U}\geq 0; r=1,\dots,s_1$$

$$\left(\sum_{\substack{j=1\\j\neq p}}^{n}\lambda_j^2\tilde{y}_{rj}^2-\varphi_r^2\tilde{y}_{rp}^2\right)_{\varepsilon_4}^{U}\geq 0; r=1,\dots,s_2$$

$$\left(\sum_{\substack{j=1\\j\neq p}}^{n}\lambda_j^1\tilde{z}_{fj}^1-\sum_{\substack{j=1\\j\neq p}}^{n}\lambda_j^2\tilde{z}_{fj}^1\right)_{\varepsilon_5}^{L}\leq 0;\quad f=1,\dots,F$$

$$\sum_{\substack{j=1\\j\neq p}}^{n}\lambda_j^1=1;$$

$$\sum_{\substack{j=1\\j\neq p}}^{n}\lambda_j^2=1;$$

$$\begin{aligned}
\theta_i^1-1 &\leq M\delta^1; & i&=1,\dots,m_1;\\
-\theta_i^1+1 &\leq M(1-\delta^1); & i&=1,\dots,m_1;\\
\theta_i^2-1 &\leq M\delta^2; & i&=1,\dots,m_2;\\
-\theta_i^2+1 &\leq M(1-\delta^2); & i&=1,\dots,m_2;\\
-\varphi_r^1+1 &\leq M\delta^1; & r&=1,\dots,s_1;\\
\varphi_r^1-1 &\leq M(1-\delta^1); & r&=1,\dots,s_1;\\
-\varphi_r^2+1 &\leq M\delta^2; & r&=1,\dots,s_2;\\
\varphi_r^2-1 &\leq M(1-\delta^2); & r&=1,\dots,s_2;
\end{aligned}$$

$$\delta^1,\delta^2 \in \{0,1\}$$

$$\theta_i^1,\varphi_r^1,\lambda_j^1,\theta_i^2,\varphi_r^2,\lambda_j^2 \geq 0; \qquad \forall i,r,j$$

Consider in the proposed model, each fuzzy number is considered as a triangular fuzzy number. So, let $\tilde{x}_{ij}^k=(x_{ij}^{kL},x_{ij}^{kM},x_{ij}^{kU})$, (k=1, 2) is a triangular fuzzy number of the $i^{th}$ input of DMU$_j$ at the $k^{th}$ stage, $\tilde{y}_{rj}^k=(y_{rj}^{kL},y_{rj}^{kM},y_{rj}^{kU})$, (k=1, 2) are the triangular fuzzy numbers of the $r^{th}$ output of DMU$_j$ at the $k^{th}$ stage. Also, $\tilde{z}_{fj}=(z_{fj}^L,z_{fj}^M,z_{fj}^U)$ is a triangular fuzzy number of the $f^{th}$ intermediate product of DMU$_j$. Also, without loss of generality, let us assume that $\varepsilon_1=\varepsilon_2=\varepsilon_3=\varepsilon_4=\varepsilon_5=\alpha$. By these transformations our model for evaluating DMU$_p$ and measuring its overall efficiency becomes as follows:

$$R_p^* = \min \frac{w_1 \left\{ \frac{1}{m_1} \sum_{i=1}^{m_1} \theta_i^1 \right\} + w_2 \left\{ \frac{1}{m_2} \sum_{i=1}^{m_2} \theta_i^2 \right\}}{w_1 \left\{ \frac{1}{s_1} \sum_{r=1}^{s_1} \varphi_r^1 \right\} + w_2 \left\{ \frac{1}{s_2} \sum_{r=1}^{s_2} \varphi_r^2 \right\}}$$

s.t.

$$\sum_{\substack{j=1 \\ j \neq p}}^{n} \lambda_j^1 (x_{ij}^{1L} + \alpha(x_{ij}^{1M} - x_{ij}^{1L})) - \theta_i^1 (x_{ip}^{1L} + \alpha(x_{ip}^{1M} - x_{ip}^{1L})) \leq 0, \qquad i=1,...,m_1;$$

$$\sum_{\substack{j=1 \\ j \neq p}}^{n} \lambda_j^2 (x_{ij}^{2L} + \alpha(x_{ij}^{2M} - x_{ij}^{2L})) - \theta_i^2 (x_{ip}^{2L} + \alpha(x_{ip}^{2M} - x_{ip}^{2L})) \leq 0, \qquad i=1,...,m_2;$$

$$\sum_{\substack{j=1 \\ j \neq p}}^{n} \lambda_j^1 (y_{rj}^{1U} - \alpha(y_{rj}^{1U} - y_{rj}^{1M})) - \varphi_r^1 (y_{rp}^{1U} - \alpha(y_{rp}^{1U} - y_{rp}^{1M})) \geq 0, \qquad r=1,...,s_1;$$

$$\sum_{\substack{j=1 \\ j \neq p}}^{n} \lambda_j^2 (y_{rj}^{2U} - \alpha(y_{rj}^{2U} - y_{rj}^{2M})) - \varphi_r^2 (y_{rp}^{2U} - \alpha(y_{rp}^{2U} - y_{rp}^{2M})) \geq 0, \qquad r=1,...,s_2;$$

$$\sum_{\substack{j=1 \\ j \neq p}}^{n} \lambda_j^1 (z_{fj}^L + \alpha(z_{fj}^M - z_{fj}^L)) - \sum_{\substack{j=1 \\ j \neq p}}^{n} \lambda_j^2 (z_{fj}^L + \alpha(z_{fj}^M - z_{fj}^L)) \leq 0, \qquad f=1,...,F; \qquad (5)$$

$$\sum_{\substack{j=1 \\ j \neq p}}^{n} \lambda_j^1 = 1; \qquad \sum_{\substack{j=1 \\ j \neq p}}^{n} \lambda_j^2 = 1;$$

$$\theta_i^1 - 1 \leq M \delta^1; \quad i=1,...,m_1; \qquad \theta_i^2 - 1 \leq M \delta^2; \quad i=1,...,m_2;$$

$$-\theta_i^1 + 1 \leq M(1-\delta^1); \quad i=1,...,m_1; \qquad -\theta_i^2 + 1 \leq M(1-\delta^2); \quad i=1,...,m_2;$$

$$-\varphi_r^1 + 1 \leq M \delta^1; \quad r=1,...,s_1; \qquad -\varphi_r^2 + 1 \leq M \delta^2; \quad r=1,...,s_2;$$

$$\varphi_r^1 - 1 \leq M(1-\delta^1); \quad r=1,...,s_1; \qquad \varphi_r^2 - 1 \leq M(1-\delta^2); \quad r=1,...,s_2;$$

$$\delta^1, \delta^2 \in \{0,1\}, \qquad \theta_i^1, \theta_i^2, \lambda_j^1, \lambda_j^2 \geq 0, \quad \forall i; \; \forall j$$

In addition, $w_1$ and $w_2$ are weights assigned to the efficiency scores in the first and second stages, respectively. Several approaches such as point allocation, paired comparisons, trade-off analysis, and regression estimates can be used to specify the weights ([37,38]). Alternatively, pairwise comparisons and eigenvalue theory proposed by [39] can be used to determine suitable weights for efficiency scores of the two stages, [9].

For each value of $\alpha \in [0,1]$ model (5) calculates the overall efficiency score for DMUp. This value is called $\alpha$-overall efficiency. Also, by using the optimal values of model (5), the $\alpha$-first stage efficiency and $\alpha$-second stage efficiency can be determined. For the purpose of integrating the obtained scores and ranking DMUs, we use the following criterion $\psi_p$ for each DMUp and is called stochastic closeness coefficient. This criterion is inspired by the closeness coefficient of TOPSIS method. Assume that *n+1* different value for $\alpha \in [0,1]$ as $\{\alpha_0, \alpha_1, ..., \alpha_n\}$ are applied to obtain the $\alpha$- efficiencies. We denoted the selected values for $\alpha$ by $\Delta$ i.e. $\Delta = \{\alpha_0, \alpha_1, ..., \alpha_n\}$. This criterion is measured as follows:

$$\psi_p = \frac{\left(\frac{\sum_{\alpha \in \Delta} R_p^\alpha}{n+1} - \min_{\alpha,j}\{R_j^\alpha\}\right)}{\left(\frac{\sum_{\alpha \in \Delta} R_p^\alpha}{n+1} - \min_{\alpha,j}\{R_j^\alpha\}\right) + \left(\max_{\alpha,j}\{R_j^\alpha\} - \frac{\sum_{\alpha \in \Delta} R_p^\alpha}{n+1}\right)}$$

In fact, $\min_{\alpha,j}\{R_j^\alpha\}$ is the worst result of $\alpha$- efficiencies among all DMUs and under all considered values for $\alpha$, so it is a kind of negative ideal value. On the other hand, the best result of $\alpha$- efficiencies among all DMUs is $\max_{\alpha,j}\{R_j^\alpha\}$, so it is a kind of positive ideal value. The idea behind the criterion $\psi_p$ is if the average obtained values for DMUp has the shortest distance from the positive ideal value and the farthest distance from the negative ideal value then DMUp should have the best ranking situation. The stochastic closeness coefficient is simply can be converted to the following relation:

$$\psi_p = \frac{\frac{\sum_{\alpha \in \Delta} R_p^\alpha}{n+1} - \min_{\alpha,j}\{R_j^\alpha\}}{\max_{\alpha,j}\{R_j^\alpha\} - \min_{\alpha,j}\{R_j^\alpha\}} \tag{6}$$

Clearly, for each p we have $0 \leq \psi_p \leq 1$. And thus we can rank DMUs according to decreasing order of the stochastic closeness coefficient of overall efficiency. By a same manner we can obtain the stochastic closeness coefficient of stage efficiencies.

## 5 Case Study

Bank Melli Iran (BMI) is the first national Iranian bank. The bank was established in 1927 by the order of the Majlis (the Iranian Parliament) and since then has consistently been one of the most influential Iranian banks. Since 1933, BMI has grown to become a large retail bank with several domestic and international branches. BMI opened its first foreign branch in Hamburg, Germany, in 1965. BMI is now the largest commercial retail bank in Iran and in the Middle East with over 3,300 branches and 43,000 employees. The aim of this paper is to evaluate 15 branches of Melli bank which are located in Hamedan province. The data are belonging to the period 2015 to 2016 and are obtained from a direct survey of the banks. As shown in Fig. 3, the two-stage performance measurement system of branches of Melli bank is comprised of two stages. Stage 1 represents the profitability and Stage 2 represents marketability banking.

*5.1 Data*

The inputs to the first stage are: ($x_1^1$) Branch costs, which consists of personnel costs, administrative costs and operating costs; ($x_2^1$) Employee, consists of the number of employees by considering their education levels; The inputs to the second stage are: ($x_1^2$) Staff, consists of the number of employees by considering their education levels; ($x_2^2$) Facilities, that consists of features such as queuing system, seating for customers, the new computer equipment and etc. The intermediate products are: ($z_1$) Deposit, which consists of short-term investment, long-term investment, and etc.; ($z_2$) Loans, which consists of Loans presented, partnership loan, loans to buy housing, and etc.; The outputs of the first stage is: ($y_1^1$) Documents, which consists of Cash and transfer documents centralized and decentralized systems, and two-fifths of the service bills in proportion to the size of their turnover. The outputs of the second stage are: ($y_1^2$) Net profit. The two-stage structure of the banking system is shown in Fig. 1.

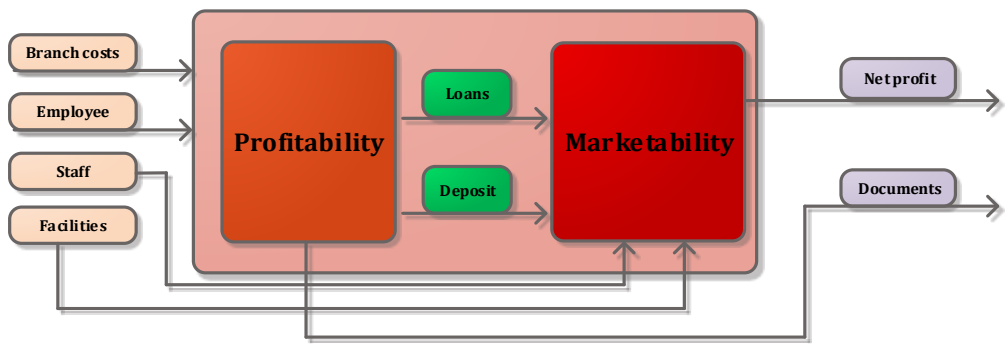

**Fig. 1:** Proposed two-stage structure of each Bank branch

To deal with the uncertainty, in this study the inputs, intermediate and outputs are considered as fuzzy numbers. Historical data of 15 branches of Melli bank (DMUs) are reported in Tables 1-3.

**Table 1:** Related fuzzy dataset for bank's Profitability (Stage 1)

| Bank branches (DMUs) | | $x_1^1$ : Branch costs | | | $x_2^1$ : Employee | | | $y_1^1$ : Documents | | |
|---|---|---|---|---|---|---|---|---|---|---|
| | | $x_1^{1L}$ | $x_1^{1M}$ | $x_1^{1U}$ | $x_2^{1L}$ | $x_2^{1M}$ | $x_2^{1U}$ | $y_1^{1L}$ | $y_1^{1M}$ | $y_1^{1U}$ |
| 1 | Emamzadeh_Abdollah | 5872332921.05 | 6053951465.00 | 7083123214.05 | 2.88 | 3.35 | 3.62 | 196329.94 | 202402.00 | 242882.40 |
| 2 | Shahrdari | 2516640000.00 | 2568000000.00 | 2901840000.00 | 3.78 | 4.20 | 4.70 | 123627.30 | 130134.00 | 143147.40 |
| 3 | Bolvar_Keshavarz | 4493097478.87 | 4937469757.00 | 5924963708.40 | 2.93 | 3.12 | 3.68 | 133654.00 | 157240.00 | 183970.80 |
| 4 | Aramgah | 13081900164.70 | 13770421226.00 | 15422871773.12 | 3.82 | 3.86 | 4.28 | 310575.72 | 316914.00 | 367620.24 |
| 5 | Bou_Ali | 15421640000.00 | 16406000000.00 | 16734120000.00 | 3.09 | 3.47 | 3.57 | 324154.60 | 334180.00 | 374281.60 |
| 6 | Dadgostari | 4930487939.52 | 5602827204.00 | 6331194740.52 | 4.33 | 4.42 | 4.46 | 152807.34 | 162561.00 | 186945.15 |
| 7 | Ghadir | 2902049607.80 | 2961275110.00 | 3405466376.50 | 4.31 | 4.68 | 5.43 | 237390.30 | 263767.00 | 316520.40 |
| 8 | Takhti | 2412386130.56 | 2566368224.00 | 2797341364.16 | 3.07 | 3.37 | 3.77 | 205789.80 | 236540.00 | 255463.20 |
| 9 | Aref | 1918832000.00 | 2231200000.00 | 2275824000.00 | 4.09 | 4.35 | 5.05 | 216897.12 | 246474.00 | 276050.88 |
| 10 | BabaTaher | 5248381794.94 | 6102769529.00 | 6896129567.77 | 3.11 | 3.46 | 3.91 | 226214.76 | 240654.00 | 257499.78 |
| 11 | Shariati | 11769377526.72 | 11888260128.00 | 12482673134.40 | 3.69 | 4.15 | 4.86 | 190521.82 | 221537.00 | 234829.22 |
| 12 | Pasdaran | 4712500732.60 | 5122283405.00 | 5378397575.25 | 2.74 | 3.15 | 3.24 | 158040.16 | 162928.00 | 177591.52 |
| 13 | Bazar | 3351772348.10 | 3943261586.00 | 4692481287.34 | 3.33 | 3.47 | 3.68 | 271714.25 | 286015.00 | 311756.35 |
| 14 | Meidan_Sepah | 4273044564.48 | 4451088088.00 | 4762664254.16 | 3.09 | 3.12 | 3.43 | 191279.70 | 212533.00 | 235911.63 |
| 15 | Meidan_Bar | 3545250408.90 | 4075000470.00 | 4482500517.00 | 3.58 | 4.02 | 4.46 | 121305.68 | 131854.00 | 141083.78 |

**Table 2:** Related fuzzy dataset for intermediate variables

| Bank branches (DMUs) | | $z_1$ : Deposit | | | $z_2$ : Loans | | |
|---|---|---|---|---|---|---|---|
| | | $z_1^L$ | $z_1^M$ | $z_1^U$ | $z_2^L$ | $z_2^M$ | $z_2^U$ |
| 1 | Emamzadeh_Abdollah | 76373055854.86 | 88805878901.00 | 92358114057.04 | 8639167893.93 | 9930078039.00 | 10128679599.78 |
| 2 | Shahrdari | 73904560975.88 | 80331044539.00 | 81134354984.39 | 10346914057.62 | 10558075569.00 | 11402721614.52 |
| 3 | Bolvar_Keshavarz | 51391897277.20 | 54096733976.00 | 63834146091.68 | 23679405346.96 | 26908415167.00 | 29061088380.36 |
| 4 | Aramgah | 149979820000.00 | 159553000000.00 | 189868070000.00 | 22845650066.11 | 23552216563.00 | 28262659875.60 |
| 5 | Bou_Ali | 166223960000.00 | 176834000000.00 | 178602340000.00 | 43398801629.10 | 48220890699.00 | 55454024303.85 |
| 6 | Dadgostari | 66461913743.45 | 78190486757.00 | 89919059770.55 | 19878215904.00 | 22086906560.00 | 23853859084.80 |
| 7 | Ghadir | 57289198855.95 | 64369886355.00 | 73381670444.70 | 25829781923.52 | 27478491408.00 | 32974189689.60 |
| 8 | Takhti | 54269850241.88 | 59637198068.00 | 66197289855.48 | 17738096057.02 | 18286696966.00 | 19749632723.28 |
| 9 | Aref | 104066810000.00 | 116929000000.00 | 122775450000.00 | 11823655263.46 | 13285005914.00 | 15543456919.38 |
| 10 | BabaTaher | 54197035389.00 | 60218928210.00 | 62025496056.30 | 15968430884.25 | 16808874615.00 | 18825939568.80 |
| 11 | Shariati | 117898800000.00 | 124104000000.00 | 132791280000.00 | 28541150072.01 | 32805919623.00 | 33133978819.23 |
| 12 | Pasdaran | 61003924555.24 | 66308613647.00 | 78907250239.93 | 16718647678.82 | 17785795403.00 | 19030801081.21 |
| 13 | Bazar | 65017789072.29 | 69911601153.00 | 73407181210.65 | 37629000836.96 | 43754652136.00 | 48567663870.96 |
| 14 | Meidan_Sepah | 40677286305.87 | 46755501501.00 | 54236381741.16 | 17701613585.63 | 19889453467.00 | 20486137071.01 |
| 15 | Meidan_Bar | 25908902079.06 | 28471320966.00 | 29894887014.30 | 30972244542.68 | 31604331166.00 | 32552461100.98 |

**Table 3:** Related fuzzy dataset for bank's Marketability (Stage 2)

| | Bank branches (DMUs) | $x_1^2$: Staff | | | $x_2^2$: Facilities | | | $y_1^2$: Net profit | | |
|---|---|---|---|---|---|---|---|---|---|---|
| | | $x_1^{2L}$ | $x_1^{2M}$ | $x_1^{2U}$ | $x_2^{2L}$ | $x_2^{2M}$ | $x_2^{2U}$ | $y_1^{2L}$ | $y_1^{2M}$ | $y_1^{2U}$ |
| 1 | Emamzadeh_Abdollah | 5.90 | 6.56 | 7.22 | 2.85 | 3.10 | 3.63 | 77143114.89 | 88670247.00 | 104630891.46 |
| 2 | Shahrdari | 6.83 | 6.90 | 8.28 | 4.02 | 4.10 | 4.88 | 1187899000.00 | 1250420000.00 | 1337949400.00 |
| 3 | Bolvar_Keshavarz | 5.83 | 6.14 | 7.37 | 1.76 | 2.00 | 2.28 | 341786012.48 | 388393196.00 | 411696787.76 |
| 4 | Aramgah | 7.51 | 7.82 | 7.98 | 4.84 | 5.15 | 6.13 | 33579521.43 | 36900573.00 | 43911681.87 |
| 5 | Bou_Ali | 6.53 | 6.87 | 7.49 | 4.47 | 5.20 | 5.82 | 1224000000.00 | 1360000000.00 | 1591200000.00 |
| 6 | Dadgostari | 7.37 | 7.44 | 8.56 | 1.83 | 1.95 | 2.30 | 45311735.56 | 48203974.00 | 57362729.06 |
| 7 | Ghadir | 7.13 | 7.75 | 8.37 | 2.82 | 2.85 | 3.11 | 332071613.30 | 338848585.00 | 345625556.70 |
| 8 | Takhti | 5.24 | 6.09 | 6.82 | 3.65 | 4.10 | 4.14 | 11177936.68 | 11891422.00 | 13080564.20 |
| 9 | Aref | 7.64 | 7.80 | 8.42 | 5.91 | 6.35 | 6.48 | 1584010000.00 | 1633000000.00 | 1926940000.00 |
| 10 | BabaTaher | 5.67 | 6.37 | 7.64 | 2.96 | 3.05 | 3.08 | 11163418.63 | 12267493.00 | 14720991.60 |
| 11 | Shariati | 6.97 | 7.11 | 7.89 | 4.47 | 4.75 | 5.23 | 28359661.33 | 31164463.00 | 32722686.15 |
| 12 | Pasdaran | 5.09 | 5.92 | 6.04 | 2.76 | 2.90 | 3.19 | 1773808763.40 | 1970898626.00 | 2325660378.68 |
| 13 | Bazar | 5.79 | 6.43 | 7.52 | 3.13 | 3.40 | 3.57 | 21250000.00 | 25000000.00 | 25250000.00 |
| 14 | Meidan_Sepah | 6.26 | 6.45 | 7.29 | 1.87 | 2.20 | 2.46 | 9353768.06 | 10278866.00 | 11101175.28 |
| 15 | Meidan_Bar | 6.19 | 7.03 | 7.38 | 1.58 | 1.65 | 1.80 | 986850000.00 | 1161000000.00 | 1184220000.00 |

*5.2 Results and Analysis*

The results of solving model (5) can be seen in Tables 4-6. These tables show the overall efficiency score and efficiency scores of the first and second stages, respectively. We ran model (5) for some values of $\alpha$ i.e. $\alpha \in \{0.0, 0.1, 0.2, 0.3, 0.4, 0.5, 0.6, 0.7, 0.8, 0.9, 1.0\}$. In this section, the efficiencies for every DMU are obtained by executing a GAMS program of model (5) at different levels of $\alpha$ for the time period 2015-2016. The model (5) is solved given $w_1$=0.5 and $w_2$=0.5. According to Table 4, we can see DMU #12 is the only DMU that is overall efficient in all values of $\alpha$, i.e. only 7% of DMUs. Also, DMU #3 is overall efficient in some values of $\alpha$. Table 6 shows better performances in the first stage. Based on Table 5, eight DMUs, i.e. 4, 5, 7, 8, 9, 12, 13, 14 are efficient in all values of $\alpha$, i.e. 53% of DMUs. Table 6 shows that the performances have been worse and only three DMUs, i.e. 3, 12 and 15, are efficient in all values of $\alpha$ in the second stage, i.e. 20% of DMUs.

From the Table 4-6 we can see there are some DMUs which are efficient in the first stage but they are inefficient in the second stage. One reason for this issue is that since these DMUs are efficient in the first stage, they produce a large amount of output and these outputs are inputs for the second stage. Thus, they consume a large amount of inputs for the second stage to produce outputs in the second stage. This point leads to decrease in their efficiency score in the second stage. Similar reason holds for inefficient DMUs in the first stage, [9].

**Table 4:** The overall efficiency of DMUs with respect to different value of $\alpha$

|    | $\alpha$=0.0 | $\alpha$=0.1 | $\alpha$=0.2 | $\alpha$=0.3 | $\alpha$=0.4 | $\alpha$=0.5 | $\alpha$=0.6 | $\alpha$=0.7 | $\alpha$=0.8 | $\alpha$=0.9 | $\alpha$=1 |
|---|---|---|---|---|---|---|---|---|---|---|---|
| 1 | 0.083 | 0.0829 | 0.0829 | 0.0828 | 0.0828 | 0.0767 | 0.0737 | 0.0731 | 0.0726 | 0.0719 | 0.0711 |
| 2 | 0.4543 | 0.4583 | 0.4623 | 0.4664 | 0.4705 | 0.4747 | 0.4789 | 0.4831 | 0.4875 | 0.4919 | 0.4964 |
| 3 | 0.9024 | 0.9107 | 0.9202 | 0.9314 | 0.9442 | 0.9591 | 0.9591 | 0.9632 | 1.029 | 1.0325 | 1.0376 |
| 4 | 0.0302 | 0.0303 | 0.0303 | 0.0304 | 0.0304 | 0.0305 | 0.0305 | 0.0305 | 0.0306 | 0.0306 | 0.0306 |
| 5 | 0.7867 | 0.7869 | 0.7863 | 0.7859 | 0.7854 | 0.785 | 0.7846 | 0.7842 | 0.7838 | 0.7835 | 0.7831 |
| 6 | 0.065 | 0.0643 | 0.0637 | 0.0631 | 0.0625 | 0.0619 | 0.0613 | 0.0607 | 0.06 | 0.0594 | 0.0588 |
| 7 | 0.2446 | 0.248 | 0.2514 | 0.2548 | 0.2584 | 0.262 | 0.2661 | 0.2705 | 0.2751 | 0.2797 | 0.2843 |
| 8 | 0.0115 | 0.0116 | 0.0116 | 0.0117 | 0.0118 | 0.0118 | 0.0119 | 0.012 | 0.0121 | 0.0122 | 0.0122 |
| 9 | 0.8048 | 0.8021 | 0.7994 | 0.7969 | 0.7944 | 0.792 | 0.7896 | 0.7874 | 0.7851 | 0.783 | 0.7808 |
| 10 | 0.0106 | 0.0106 | 0.0106 | 0.0106 | 0.0106 | 0.0105 | 0.0105 | 0.0105 | 0.0105 | 0.0105 | 0.0105 |
| 11 | 0.0167 | 0.0169 | 0.0172 | 0.0174 | 0.0177 | 0.0179 | 0.0182 | 0.0185 | 0.0187 | 0.019 | 0.0193 |
| 12 | 1.3633 | 1.3558 | 1.3479 | 1.3387 | 1.3296 | 1.3205 | 1.3115 | 1.3024 | 1.2933 | 1.1607 | 1.0227 |
| 13 | 0.0202 | 0.0205 | 0.0208 | 0.0212 | 0.0215 | 0.0218 | 0.0222 | 0.0225 | 0.0229 | 0.0233 | 0.0237 |
| 14 | 0.0152 | 0.015 | 0.0149 | 0.0147 | 0.0146 | 0.0145 | 0.0144 | 0.0143 | 0.0142 | 0.0149 | 0.0155 |
| 15 | 0.6878 | 0.6903 | 0.6926 | 0.6938 | 0.6949 | 0.6961 | 0.6973 | 0.6985 | 0.6998 | 0.7011 | 0.7024 |

**Table 5:** The first stage efficiency of DMUs with respect to different value of $\alpha$

|    | $\alpha$=0.0 | $\alpha$=0.1 | $\alpha$=0.2 | $\alpha$=0.3 | $\alpha$=0.4 | $\alpha$=0.5 | $\alpha$=0.6 | $\alpha$=0.7 | $\alpha$=0.8 | $\alpha$=0.9 | $\alpha$=1 |
|---|---|---|---|---|---|---|---|---|---|---|---|
| 1 | 1.1617 | 1.1399 | 1.1168 | 1.0932 | 1.069 | 0.8269 | 0.7763 | 0.7449 | 0.7124 | 0.6489 | 0.6249 |
| 2 | 0.4933 | 0.4937 | 0.4942 | 0.4946 | 0.4949 | 0.4952 | 0.4955 | 0.4958 | 0.4961 | 0.4962 | 0.4964 |
| 3 | 0.7254 | 0.7429 | 0.7626 | 0.785 | 0.8105 | 0.858 | 0.8585 | 0.8654 | 1.0052 | 1.0062 | 1.0087 |
| 4 | 1.3623 | 1.3619 | 1.3606 | 1.3536 | 1.3464 | 1.3389 | 1.3313 | 1.3233 | 1.3152 | 1.3069 | 1.2981 |
| 5 | 1.4338 | 1.4182 | 1.412 | 1.4057 | 1.3994 | 1.3931 | 1.3866 | 1.3802 | 1.3737 | 1.3671 | 1.3604 |
| 6 | 0.7758 | 0.7738 | 0.7718 | 0.7699 | 0.768 | 0.7662 | 0.7644 | 0.7628 | 0.7612 | 0.7597 | 0.7583 |

| 7 | 1.2221 | 1.2113 | 1.2003 | 1.1891 | 1.1776 | 1.166 | 1.1537 | 1.1407 | 1.1276 | 1.1141 | 1.1004 |
| 8 | 2.1244 | 2.1285 | 2.1325 | 2.1366 | 2.1407 | 2.1448 | 2.1489 | 2.153 | 2.1572 | 2.1613 | 2.1655 |
| 9 | 1.2196 | 1.2089 | 1.1984 | 1.1882 | 1.178 | 1.1681 | 1.1583 | 1.1486 | 1.139 | 1.1296 | 1.1203 |
| 10 | 0.7413 | 0.7407 | 0.7403 | 0.7401 | 0.7402 | 0.7406 | 0.7413 | 0.7423 | 0.7436 | 0.7452 | 0.7462 |
| 11 | 0.4918 | 0.4922 | 0.4926 | 0.5147 | 0.5127 | 0.5108 | 0.5089 | 0.5071 | 0.5054 | 0.5037 | 0.5021 |
| 12 | 1.0332 | 1.0296 | 1.0259 | 1.0217 | 1.0176 | 1.0136 | 1.0099 | 1.0062 | 1.0026 | 0.8412 | 0.6926 |
| 13 | 1.1866 | 1.1903 | 1.1899 | 1.1894 | 1.1899 | 1.1901 | 1.19 | 1.1896 | 1.1888 | 1.1879 | 1.1866 |
| 14 | 1.6443 | 1.6372 | 1.6304 | 1.6237 | 1.6172 | 1.3963 | 1.354 | 1.3373 | 1.3201 | 1.2952 | 1.275 |
| 15 | 0.4363 | 0.4356 | 0.4216 | 0.4202 | 0.4188 | 0.4174 | 0.4162 | 0.4149 | 0.4137 | 0.4126 | 0.4115 |

**Table 6:** The second stage efficiency of DMUs with respect to different value of $\alpha$

|  | $\alpha=0.0$ | $\alpha=0.1$ | $\alpha=0.2$ | $\alpha=0.3$ | $\alpha=0.4$ | $\alpha=0.5$ | $\alpha=0.6$ | $\alpha=0.7$ | $\alpha=0.8$ | $\alpha=0.9$ | $\alpha=1$ |
|---|---|---|---|---|---|---|---|---|---|---|---|
| 1 | 0.0412 | 0.0412 | 0.0412 | 0.0412 | 0.0413 | 0.0425 | 0.0418 | 0.0418 | 0.0418 | 0.0422 | 0.042 |
| 2 | 0.4136 | 0.421 | 0.4285 | 0.4362 | 0.4441 | 0.4522 | 0.4604 | 0.4689 | 0.4778 | 0.487 | 0.4965 |
| 3 | 1.1164 | 1.1114 | 1.1066 | 1.1021 | 1.0978 | 1.076 | 1.0754 | 1.0754 | 1.0527 | 1.0526 | 1.0526 |
| 4 | 0.0118 | 0.0118 | 0.0119 | 0.012 | 0.012 | 0.0121 | 0.0121 | 0.0122 | 0.0123 | 0.0123 | 0.0124 |
| 5 | 0.4779 | 0.4788 | 0.4799 | 0.4809 | 0.482 | 0.4832 | 0.4844 | 0.4857 | 0.487 | 0.4883 | 0.4897 |
| 6 | 0.0364 | 0.0361 | 0.0359 | 0.0357 | 0.0354 | 0.0352 | 0.035 | 0.0347 | 0.0345 | 0.0342 | 0.0339 |
| 7 | 0.1258 | 0.1282 | 0.1306 | 0.1332 | 0.1358 | 0.1385 | 0.1414 | 0.1446 | 0.1479 | 0.1512 | 0.1547 |
| 8 | 0.0049 | 0.0049 | 0.0049 | 0.0049 | 0.0049 | 0.0049 | 0.005 | 0.005 | 0.005 | 0.005 | 0.0051 |
| 9 | 0.4829 | 0.4855 | 0.4881 | 0.4908 | 0.4935 | 0.4961 | 0.4988 | 0.5015 | 0.5041 | 0.5068 | 0.5095 |
| 10 | 0.0059 | 0.0059 | 0.0059 | 0.0059 | 0.0059 | 0.0059 | 0.0059 | 0.0059 | 0.0059 | 0.0059 | 0.0059 |
| 11 | 0.0095 | 0.0097 | 0.0098 | 0.01 | 0.0102 | 0.0104 | 0.0106 | 0.0108 | 0.011 | 0.0112 | 0.0114 |
| 12 | 1.9318 | 1.9145 | 1.8811 | 1.8593 | 1.8375 | 1.8157 | 1.7936 | 1.7714 | 1.749 | 1.7243 | 1.7021 |
| 13 | 0.0096 | 0.0097 | 0.0099 | 0.01 | 0.0102 | 0.0103 | 0.0105 | 0.0107 | 0.0109 | 0.0111 | 0.0112 |
| 14 | 0.0074 | 0.0073 | 0.0072 | 0.0071 | 0.0071 | 0.007 | 0.007 | 0.0069 | 0.0069 | 0.0072 | 0.0076 |
| 15 | 1.1858 | 1.1916 | 1.1987 | 1.2041 | 1.2095 | 1.2148 | 1.2201 | 1.2254 | 1.2306 | 1.2358 | 1.2411 |

To better comparison, these results are provided in Fig. 2. This figure shows the efficiency scores of bank branches at the different levels of $\alpha$ in an integrated form. According to these tables and figure, it seems that, the performance of DMUs in the first stage is relatively better than second stage. From Fig. 2 it is clear that DMU #8, i.e. "Takhti" has been recognized for its best performance in the first stage in the different levels of $\alpha$ (It has maximum efficiency in different values of $\alpha$). Also, we can see DMU #12, i.e. "Pasdaran" has been recognized for its best performance in the second stage in the different levels of $\alpha$ and because of big difference between its efficiency and others' efficiencies, this DMU has been recognized as the best overall performance, (Fig. 2a).

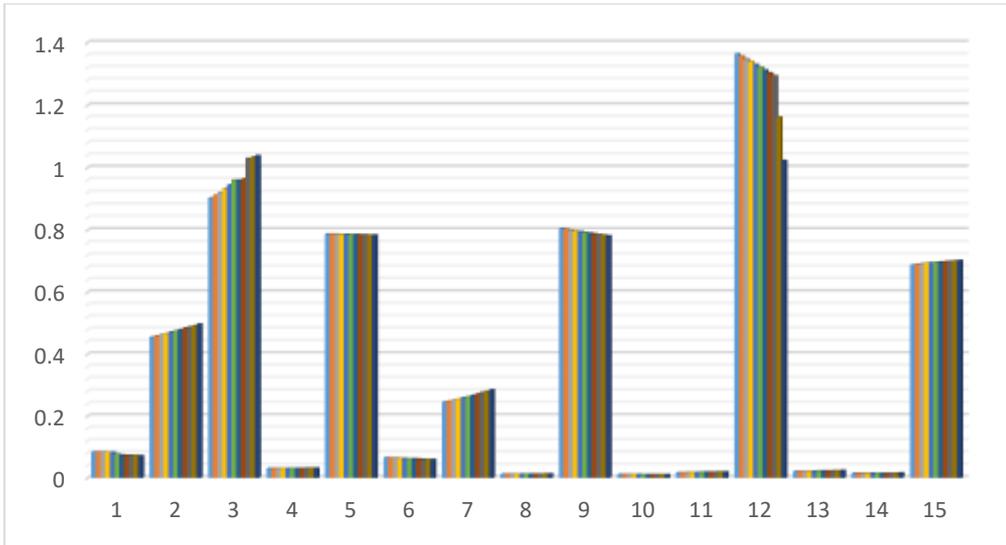

(a) Overall efficiency

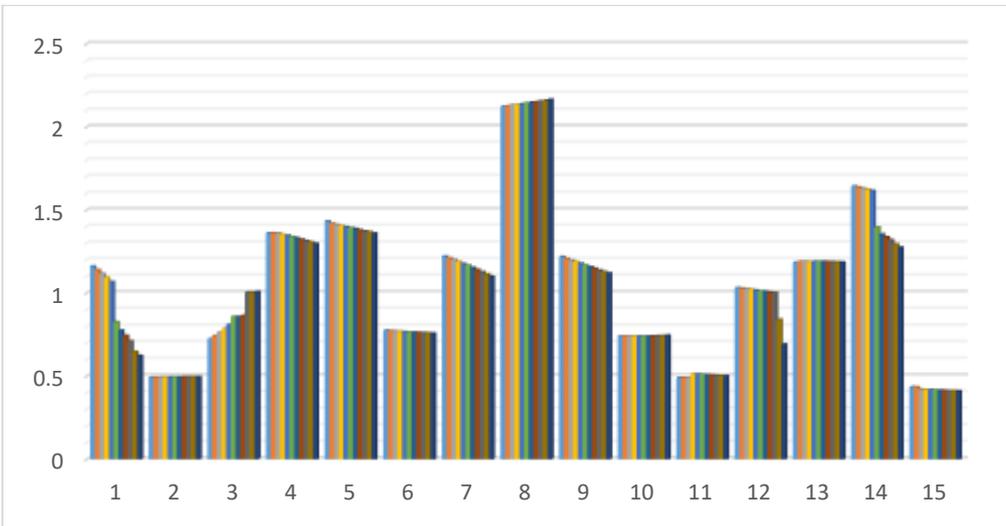

(b) First stage efficiency

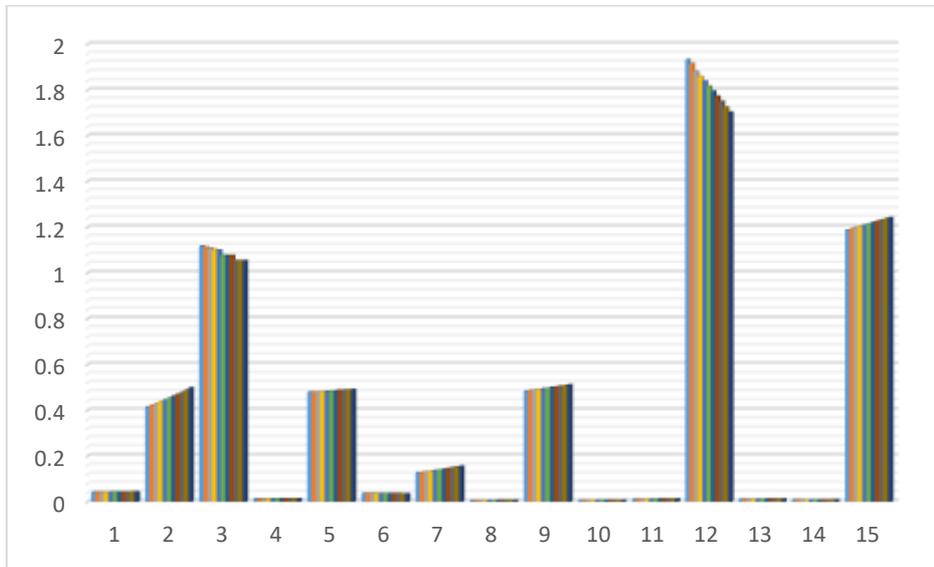

(c) Second stage efficiency

**Fig. 2:** Various efficiency scores at different levels of $\alpha$

The values of the stochastic closeness coefficient for overall, first stage and second stage efficiencies alongside rankings are shown in Table 7. From Table 7 we can see the DMU #12 (Pasdaran) and DMU #10 (BabaTaher) have the best and the worst overall performances among all DMUs, respectively. A slight note to the last row of Table 7 shows that in average the performances of DMUs in the first stage are better than their performances in the second stages. This result had been stated from Fig. 2, too.

**Table 7:** Stochastic closeness coefficients (SCCs) and related rankings

| Bank Branches (DMUs) | Overall | | First Stage | | Second Stage | |
|---|---|---|---|---|---|---|
| | SCC | Rank | SCC | Rank | SCC | Rank |
| **Emamzadeh_Abdollah** | 0.049594108 | 8 | 0.279853 | 9 | 0.019074 | 8 |
| **Shahrdari** | 0.343314876 | 6 | 0.04775 | 14 | 0.232701 | 6 |
| **Bolvar_Keshavarz** | 0.691044836 | 2 | 0.238543 | 10 | 0.559783 | 3 |
| **Aramgah** | 0.014743831 | 10 | 0.528295 | 4 | 0.003727 | 10 |
| **Bou_Ali** | 0.572543143 | 4 | 0.561103 | 3 | 0.248345 | 5 |
| **Dadgostari** | 0.037981829 | 9 | 0.202832 | 11 | 0.015715 | 9 |
| **Ghadir** | 0.186777593 | 7 | 0.429845 | 7 | 0.06973 | 7 |
| **Takhti** | 0.00100129 | 14 | 0.990262 | 1 | 2.83E-05 | 15 |
| **Aref** | 0.577925918 | 3 | 0.432654 | 6 | 0.254941 | 4 |
| **BabaTaher** | 3.36003E-05 | 15 | 0.188804 | 12 | 0.000519 | 14 |
| **Shariati** | 0.005510456 | 12 | 0.052741 | 13 | 0.002864 | 11 |
| **Pasdaran** | 0.942886135 | 1 | 0.320322 | 8 | 0.940106 | 1 |
| **Bazar** | 0.008406806 | 11 | 0.444189 | 5 | 0.00284 | 12 |
| **Meidan_Sepah** | 0.003138272 | 13 | 0.602678 | 2 | 0.00117 | 13 |
| **Meidan_Bar** | 0.506632708 | 5 | 0.004794 | 15 | 0.62765 | 2 |
| **Average** | 0.263622923 | | 0.356047103 | | 0.198613 | |

From Table 7, we can conclude that 47% of branches have efficiency more than average in Stage 1 and that 40% of branches have efficiency more than average in Stage 2. The other notable result of Table 8 is to see there are eight bank branches that are recognized as the efficient branches in the first stage in all values of $\alpha$, which means there is no efficiency loss during operating. Therefore, the branches in profitability stage have compromise performances. Also, there are three bank branches that are recognized as the efficient branches in the second stage in all values of $\alpha$. Therefore, the branches in marketability stage do not have compromise performances. But, only there is one bank branch, which is "Pasdaran" branch, that worked efficiently in both stage one and two in all values of $\alpha$, and therefore, is recognized as the only overall efficient branch among these 15 branches.

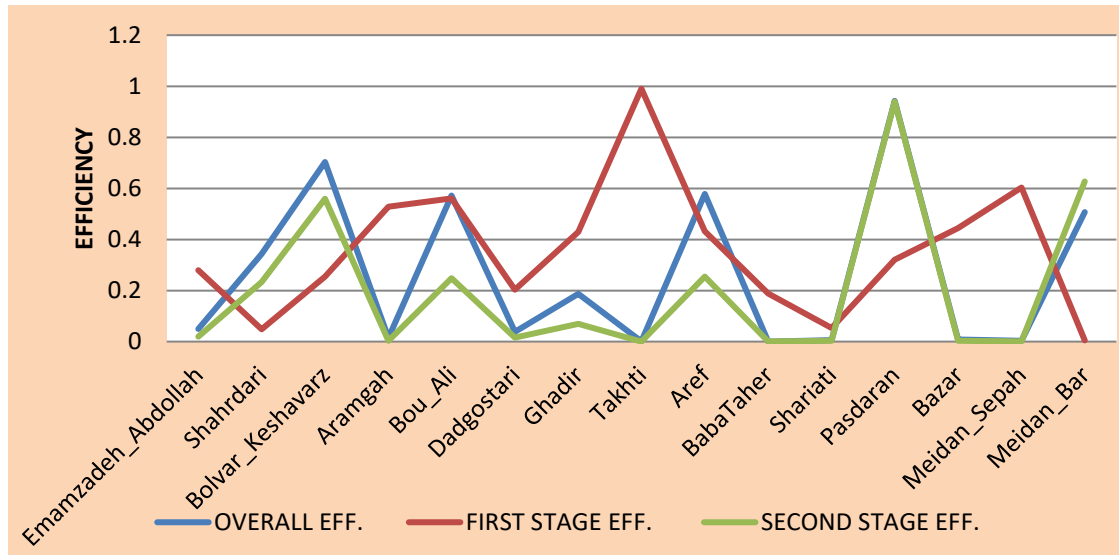

**Fig. 3:** Comparison among all stochastic closeness coefficients

Refer to Fig. 3 for an illustrative comparison among the results. Fig. 3 illustrates the graphical representations of the stochastic closeness coefficient results using fuzzy input–intermediate-output data; and it is clear that the first stage efficiency score of DMU "Takhti" is the highest value among all efficiency scores and the next place is for DMU "Pasdaran" for its second stage efficiency score. Finally, the poor performance of DMU "BabaTaher" in all situations is quite clear.

### *5.3 Comparison with other methods*

By examining the literature, it can be seen that there are a number of articles on the evaluation of decision-making units with fuzzy two-stage DEA models. Table 8 provides a comparison among the proposed model and some of the existing fuzzy two-stage DEA models. For this comparison, some important criteria such as novelty in two-stage model, non-parametric form, ability to efficiency decomposition, providing an integrated index and application in banking industry are considered.

**Table 8:** Comparison among fuzzy two-stage DEA models

|  | Novel Network Model | Parametric/ Non-Parametric | Efficiency Decomposition | integrated index | Applied in Banking Industry |
|---|---|---|---|---|---|
| **Tavana et al. [40]** | √ | Parametric | √ | × | × |
| **Shermeh et al. [41]** | √ | Parametric | √ | × | × |
| **Liu [42]** | × | Parametric | √ | × | × |
| **Soltanzadeh and Omrani [43]** | √ | Parametric | × | × | × |
| **Simsek and Tüysüz [44]** | √ | Parametric | × | × | × |
| **Hatami-Marbini and Saati [45]** | √ | Parametric | × | × | × |
| **Zhou et al. [46]** | √ | Parametric | × | × | × |
| **Tavana and Khalili-Damghani [24]** | √ | Parametric | √ | × | √ |
| **Tavana et al. [47]** | √ | Non-Parametric | √ | × | × |
| **Proposed Model** | √ | Non-Parametric | √ | √ | √ |

According to Table 8, the method proposed in this paper has several advantages over the existing models. On the other hand, the method proposed in this study is more general compared with the existing models. The non-parametric nature of the proposed model is a very important feature that makes the evaluation and ranking not dependent on $\alpha$ values. Additionally, the proposed model provides an integrated index that helps decision maker to make robust decisions.

*5.4 Sensitivity analysis on stages' weights*

This study is done by considering the equal values for weights of stages. In this section, we check the importance of weight change and its effect on the final ranking of DMUs. For this purpose, we consider four extra cases for weights as Table 9. These four cases have been selected due to consider the large difference as well as the small difference between the weights.

**Table 9:** Extra cases for stages' weights

|  | $w_1$ | $w_2$ |
|---|---|---|
| **Case 1** | 0.15 | 0.85 |
| **Case 2** | 0.45 | 0.55 |
| **Case 3** | 0.55 | 0.45 |
| **Case 4** | 0.85 | 0.15 |
| **Main Case** | 0.5 | 0.5 |

The proposed Model is run for all cases and the obtained rankings are illustrated in Table 10. The case of equal weights is called the main case in these tables. In order to check the relation among the results we employ the Spearman's correlation coefficient.

**Table 10**: The results of rankings for all cases

| No. | DMU | Main Case | CASE 1 | CASE 2 | CASE 3 | CASE 4 | Sensitivity |
|---|---|---|---|---|---|---|---|
| 1 | Emamzadeh_Abdollah | 8 | 8 | 8 | 8 | 8 | 0.000 |
| 2 | Shahrdari | 6 | 6 | 6 | 6 | 6 | 0.000 |
| 3 | Bolvar_Keshavarz | 2 | 3 | 5 | 5 | 4 | 1.304 |
| 4 | Aramgah | 10 | 10 | 10 | 10 | 10 | 0.000 |
| 5 | Bou_Ali | 4 | 5 | 4 | 2 | 1 | 1.643 |
| 6 | Dadgostari | 9 | 9 | 9 | 9 | 9 | 0.000 |
| 7 | Ghadir | 7 | 7 | 7 | 7 | 5 | 0.894 |
| 8 | Takhti | 14 | 15 | 14 | 14 | 14 | 0.447 |
| 9 | Aref | 3 | 4 | 2 | 3 | 2 | 0.837 |
| 10 | BabaTaher | 15 | 14 | 15 | 15 | 15 | 0.447 |
| 11 | Shariati | 12 | 12 | 12 | 12 | 13 | 0.447 |
| 12 | Pasdaran | 1 | 1 | 1 | 1 | 3 | 0.894 |
| 13 | Bazar | 11 | 11 | 11 | 11 | 11 | 0.000 |
| 14 | Meidan_Sepah | 13 | 13 | 13 | 13 | 12 | 0.447 |
| 15 | Meidan_Bar | 5 | 2 | 3 | 4 | 7 | 1.924 |
| **Spearman's correlation coefficient** | | | 0.975 | 0.975 | 0.975 | 0.95 | |

The Spearman's correlation coefficient among the results of obtained rankings of the current case and the new extra four cases are calculated as the last row of Table 10. The results indicate that, there is a high and meaningful correlation between obtained results. This fact indicate the high similarity among the obtained rankings. This fact shows that the weight change has a low effect on the final ranking of DMUs. However, the rankings are not completely the same and there are some differences among the results. Last column of Table 10 shows the sensitivity of DMUs with respect to weight changes. We can see that five DMUs, i.e. {1, 2, 4, 6 and 13} are not sensitive to weight change and the other DMUs are very little sensitive to weight changes. DMU 15 is the most sensitive DMU among all.

### *5.5 Analysis and further discussions*

Now, by neglecting the intermediate products, we consider the DMUs as black boxes and measure the sustainability of suppliers. It is shown that in such cases, using the conventional DEA models may lead to the biased results. For instance, the conventional DEA models cannot specify the inefficiency reasons in network structured DMUs ([48]; [24]; [37,49]). Also, there are several studies that show the deficiency of traditional DEA models (e.g., [50], [51], [52], [53,54]).

Unlike the traditional DEA models, the two-stage DEA model can examine the structure and processes within DMUs. This helps managers to identify the inefficiency sources within DMUs ([55]; [56]; [57]; [58]). To investigate the effects of intermediate products on DMUs' performance, we assess the performance of branches of BMI assuming there is no intermediate measure. To this end, the Branch costs, Employee and Staff, Facilities are considered as inputs. The Documents and Net profit are considered as outputs. To compare the overall sustainability scores of the two-stage DEA model and the "black box", the following fuzzy single-stage DEA model is formulated:

$$\min \frac{\frac{1}{m_1}\sum_{i=1}^{m_1} \theta_i^1}{\frac{1}{s_1}\sum_{r=1}^{s_1} \phi_r^1}$$

s.t.

$$\sum_{\substack{j=1 \\ j \neq p}}^{n} \lambda_j^1 (x_{ij}^{1L} + \alpha(x_{ij}^{1M} - x_{ij}^{1L})) - \theta_i^1 \left(x_{ip}^{1L} + \alpha(x_{ip}^{1M} - x_{ip}^{1L})\right) \leq 0; i = 1, \dots, m_1$$

$$\sum_{\substack{j=1 \\ j \neq p}}^{n} \lambda_j^1 \left(y_{rj}^{1U} - \alpha(y_{rj}^{1U} - y_{rj}^{1M})\right) - \phi_r^1 \left(y_{rp}^{1U} - \alpha\left(y_{rp}^{1U} - y_{rp}^{1M}\right)\right) \geq 0; r = 1, \dots, s_1$$

(7)

$$\sum_{\substack{j=1 \\ j \neq p}}^{n} \lambda_j^1 = 1;$$

$$\theta_i^1 - 1 \leq M\delta^1; \quad i = 1, \dots, m_1;$$
$$-\theta_i^1 + 1 \leq M(1 - \delta^1); \quad i = 1, \dots, m_1;$$
$$-\varphi_r^1 + 1 \leq M\delta^1; \quad r = 1, \dots, s_1;$$
$$\varphi_r^1 - 1 \leq M(1 - \delta^1); \quad r = 1, \dots, s_1;$$
$$\delta^1 \in \{0,1\}$$
$$\theta_i^1, \varphi_r^1, \lambda_j^1 \geq 0; \quad \forall i, r, j$$

The result of stochastic closeness coefficient (6) are shown in Table 11. Based on Table 11, the average of obtained scores by model (7) is more than the fuzzy two-stage DEA model. This fact indicates that there are some inefficiencies that the single-stage model cannot recognize them.

**Table 11:** Black-box results

|  | Main | Black Box |
|---|---|---|
| 1 | 0.04959 | 0.55989 |
| 2 | 0.34331 | 0.10704 |
| 3 | 0.70385 | 0.43743 |
| 4 | 0.01474 | 0.71114 |
| 5 | 0.57254 | 0.93810 |
| 6 | 0.03798 | 0.02485 |
| 7 | 0.18678 | 0.77245 |
| 8 | 0.00100 | 0.84774 |
| 9 | 0.57793 | 0.95568 |
| 10 | 0.00003 | 0.39495 |
| 11 | 0.00551 | 0.05241 |
| 12 | 0.94289 | 0.57438 |
| 13 | 0.00841 | 0.96168 |
| 14 | 0.00314 | 0.59329 |
| 15 | 0.50663 | 0.00959 |
| Average | 0.26362 | 0.52938 |

The results of Table 11 show that the overall performance of DMUs, in their single stage form is very far from the results obtained by the proposed two-stage DEA model. The differences imply that the black-box DEA model cannot properly represent the overall performance of bank branches. The results show that if the intermediate factors are not involved in the evaluation process, there might be biased results.

## *5.6 Managerial Insights*

Not only performance measurement in bank branches is essential but it also plays a critical role as an element of productive banking industry operations on strategic and operational levels. Mathematical models and analytic approaches are powerful tools in the performance measurement of bank branches and they enable managers to obtain helpful information to inform strategic and operational decisions. One of the popular and rigorous approaches used by managers to evaluate the efficiency in banking industry is DEA model. The findings of this study can also increase operations managers' confidence in the right decision-making for performance measurement of bank branches.

Classical DEA models consider each DMU as a black box and do not care about internal structure of DMUs. Also, primary DEA models assume that data are known exactly. However, in real world, there might be stochastic data. Our proposed two-stage DEA model can evaluate the bank branches in presence of fuzzy data. Moreover, the feature of using two-stage structure can help managers to identify any inefficient resources in each stage of a banking operation and address these by making the right decisions. Another issue is that since the initial investment in banking industries under uncertain conditions can be costly, time-consuming and risky, powerful performance measurement techniques, including the fuzzy two-stage DEA model presented in this study, can serve as appropriate decision support system tools.

In the case study section, the proposed model has been applied to evaluate the efficiency of 15 bank branches in Hamedan. According to the derived results, only one bank branch was recognized as efficient during the examined period 2014-2015 at all significant levels. This fact provides managers with information about which branches need to be actively developed so as to trigger innovation and growth. It also allows managers to identify productive investment and appropriate management activities. In the first stage, the sub-process of profitability measurement, eight branches were identified as efficient (at all significant levels). In the second stage, the sub-process of marketability measurement, three branches were recognized to perform efficiently (at all significant levels). From a statistical viewpoint, the efficiency of the first stage must be higher than that of the second stage. This indicates that the low efficiency scores obtained for the two-stage processes were mainly due to the low efficiency scores of the corresponding second stages, that is, the marketability efficiency values.

## 6 Conclusion

The assessment of the banks' performance has always been of interest due to their crucial role in most economic activities and the maintenance of the health of monetary markets and economic conditions. Data envelopment analysis has been found to be a well-known methodology for measuring performance of bank branches. However, conventional DEA model makes difficult, if not impossible, to understand what are the sub-processes and the interactions causing the inefficiency of a DMU. In addition, in the banking environment, there is always a need for tools that allow one to uncover the inefficiencies that can affect the different components of an operation. Moreover, many banking operations take the form of two-stage processes. In this paper, we considered a novel two-stage DEA model based on the modified version of ERM method. The aim of this study was evaluation the 15 branches of Melli bank in Hamedan province. After determining the input, intermediate and output variables, it is realized that generally they weren't precisely known and as a result they couldn't be considered in exact form. For this reason, the data was stated as fuzzy data. We used triangular fuzzy data to state the complexity of data. Therefore, we further extended our proposed two-stage DEA model to deal with fuzzy data. After that we presented a method for solving the proposed fuzzy DEA model based on the concept of alpha cut and possibility approach.

In the case study section, the proposed model was applied to evaluate the efficiency of 15 bank branches in Hamedan. The proposed fuzzy two-stage DEA model was coded using GAMS 23.6 software and was conducted by considering some different values of $\alpha$. The obtained values were integrated by using the proposed stochastic closeness coefficient. Based on the results of the proposed stochastic closeness coefficient the best and the worst performances were determined. According to the derived results, 14 out of the 15 bank branches analyzed turned out to be overall inefficient. Only one bank branch was recognized as efficient during the examined period 2014-2015 at all significant levels.

We conclude with a few possible research directions towards which to extend the results of this study. Our approach could be extended to consider other kinds of data such as dual-role data, stochastic data and so on. In this paper, we applied our proposed model for evaluating the performance of bank branches. It seems that our proposed model can be used in other problems such as evaluating the sustainability of suppliers, regional R&D processing, evaluating non-life insurance companies, efficiency evaluation of production lines, efficiency evaluation of hospitals which have many wards interacting with each other and have network structure, and so on. Developing a fuzzy dynamic two-stage DEA model will be another interesting research topic. The proposed model may be extended to cases where the intermediate products could be lost or added from external sources.